\documentclass[24 pt]{amsart}

\usepackage{booktabs} 
\usepackage{amssymb,color}
\usepackage{latexsym, algorithm, algpseudocode}
\usepackage{graphicx}
\usepackage{amsmath}
\usepackage[english]{babel}
\usepackage{amsfonts}
\usepackage[square,comma,sort&compress]{natbib}

\allowdisplaybreaks
\iftrue 
\usepackage{amsmath}
\usepackage{amsfonts}
\newcommand{\field}[1]{\mathbb{#1}}
\DeclareMathOperator{\PR}{\field{P}}             
\DeclareMathOperator{\E}{\field{E}}              
\def\N{\field{N}}                                
\def\R{\field{R}}                                
\def\F{\field{F}}                                

\else
\def\PR{\mathop{\rm I\kern -0.20em P}\nolimits}  
\def\E{\mathop{\rm I\kern -0.20em E}\nolimits}   
\def\N{\mathop{\rm I\kern -0.20em N}\nolimits}   
\def\R{\mathop{\rm I\kern -0.20em R}\nolimits}   
\def\F{\mathop{\rm I\kern -0.20em F}\nolimits}   
\fi

\newtheorem{thm}{Theorem}[section]

\newtheorem{prop}[thm]{Proposition}

\numberwithin{equation}{section}

\title[Robust counterfactual inferences using Feature Learning]{Robust counterfactual inferences \\ using Feature learning and their applications } 
\author[A. Mitra, K. Achan and S. Kumar]{Abhimanyu Mitra, Kannan Achan and Sushant Kumar}
\address{Abhimanyu Mitra,
Walmart Labs\\
600 West California Avenue\\
Sunnyvale, CA.}
\email{AMitra@walmartlabs.com}
\address{Kannan Achan,
Walmart Labs\\
600 West California Avenue\\
Sunnyvale, CA.}
\email{KAchan@walmartlabs.com}
\address{Sushant Kumar,
Walmart Labs\\
600 West California Avenue\\
Sunnyvale, CA.}
\email{SKumar4@walmartlabs.com}

\normalsize

\begin{document}

\maketitle

\begin{abstract}
In a wide variety of applications, including personalization, we want to measure the difference in outcome due to an intervention and thus have to deal with counterfactual inference. The feedback from a customer in any of these situations is only ``bandit feedback'' - that is, a partial feedback based on whether we chose to intervene or not. Typically randomized experiments are carried out to understand whether an intervention is overall better than no intervention. Here we present a feature learning algorithm to learn from a randomized experiment where the intervention in consideration is most effective and where it is least effective rather than only focusing on the overall impact, thus adding a context to our learning mechanism and extract more information. From the randomized experiment, we learn the feature representations which divide the population into subpopulations where we observe statistically significant difference in average customer feedback between those who were subjected to the intervention and those who were not, with a level of significance $l$, where $l$ is a configurable parameter in our model. We use this information to derive the value of the intervention in consideration for each instance in the population. With experiments, we show that using this additional learning, in future interventions, the context for each instance could be leveraged to decide whether to intervene or not. 
\end{abstract}

\section{Introduction}

One of the most common form of data related to a Web service is customer feedback available in some form of interaction log of customers, when they are interacting with the Web service. However, typically these interaction logs are only partial information, also known as ``bandit feedback'', as it is contingent upon the prediction made by the system about what is the best way to present the service or which service to present to the customer; see \cite{swaminathan:joachims:2015}. For example, in personalization, the prediction is about adapting the content according to the customer, which most often turns out to be picking appropriate contents from a content pool based on customer features like the past browse and purchase activity of the customer. In many of the situations, we want to know how an alternate system of making predictions would have performed which brings us to the realm of counterfactual inference. For example, in personalization, an alternate system could be a different method of picking appropriate contents from the content pool based on customer features.

The problem of counterfactual inference has a rich literature with some the earlier works dating back to the 1970s and some of the latest appearing in the last few years; see for example \cite{lewis:1973, rubin:1974, rosenbaum:rubin:1983, rubin:2005, bang:robins:2005, laan:petersen:2007, hill:2011, dudik:langford:li:2011, austin:2011, chernozhukov:frenandezval:melly:2013, bottou:others:2013, swaminathan:joachims:2015, johansson:shalit:sontag:2016}. However, most of this literature is focussed on observational studies rather than a controlled experiment and dealing with the challenge of eliminating selection bias from the inference. Randomized experiments are known to eliminate this selection bias, but conducting a randomized experiment is costly and it might be even impossible to conduct one such in certain observational studies. Some recent research has been devoted to clever experimental designs which could lead to reducing the cost of the randomized experiment; see \cite{kohavi:longbotham:sommerfield:henne:2009, tang:agarwal:obrien:meyer:2010, bottou:others:2013, johnson:lewis:nubbermeyer:2017}. In our research, we focus on none of the above problems and accept as our base a randomized experiment framework, where we choose one system as the incumbent and make interventions in that system in randomly chosen situations by overwriting the incumbent system with the predictions of the new system. Thus our base framework already incurs the cost associated with a randomized experiment and therefore for us, a simple comparison of average customer feedbacks (click-through-rates, sales revenue per impression etc.) for the interventions with those where the incumbent system is not intervened, would yield which system is performing better, free of any selection bias. Therefore, we also do not need sophisticated measurement techniques for counterfactual inference as  is required in the observational studies to eliminate selection bias. However, even if we do not have a selection bias, we still might get an inconclusive result due to the noise, when the difference in average customer feedback between those subjected to the intervention and those who were not, is not statistically significantly different from $0$, with a level of significance $l$. Note that the noise is a result of the difference in feedbacks from different instances, mostly due to different preferences as well as different inclinations to provide feedback (some customers are more likely to click than others irrespective of the level of satisfaction with the service). So, we ask the question: what if we group together some of the instances with very similar preferences and very similar inclinations to provide feedback, so that we could get a conclusive result for that group? While a conclusive result for a particular group might not necessarily mean a conclusive global result (it could still be inconclusive at the global level due to noise), we can utilize this information in future system design. In other words, having incurred the cost of a randomized experiment, we ask if we can extract more information from the experiment rather than a global comparison between the two systems? More specifically, we ask the question: can the context of an instance indicate in a robust way (robust against the inherent noise in customer feedback) which system is more suitable rather than picking a system to be used globally (the system which overall performs better)? 

We propose a more personalized approach to learn a system's performance. While one system, say system A, might be overall better than another system, say system B, there might be instances where it is better to predict using system B. The context of an instance might guide us in predicting whether the instance will prefer predictions from system A or predictions from system B. For example, in personalization, we may not yet have found a method of personalizing contents that, based on user features, decides appropriate contents from a content pool, and is universally better than all other methods of personalization. A more realistic scenario is where we have a pool of methods, each of which is best for some considerably large subpopulation, but none of them universally dominates all the others.  In such a case, a personalization system which lets the methods divide and conquer, will perform much better than a personalization system which chooses only one of them. A personalization system that lets the individual personalization methods divide and conquer, will work in two layers, where in the first layer of personalization, based on user features (context), the system decides on a method and then, using the chosen method and user features, pick appropriate contents from a content pool for the user. The relative success of this two-layered prediction method when compared to picking the overall better-performing system to be used globally, depends on whether there is enough dissidence about the system preference among the instances. This property of the two-layered prediction is similar to personalization itself, which, compared to a global method of picking best content, works best when the content preferences are vastly different for different instances.

However, finding the set of instances which might prefer a different system than the rest is a combinatorial challenge as the number of subsets explode quickly. Also, unless the subsets are characterized by a function of the contexts of the constituent instances, we could not make the learning useful in future system design. Since the context is usually a feature vector of several dimensions (for example, in user-based personalization, a user's past browsing and purchase history could become the context), characterizing the subsets of instances where the constituent instances prefer a system that is different than those outside the subset, in terms of the contexts of the constituent instances, is impossible to achieve by iteratively checking each possible function of the contexts of the instances to define each possible subset. 

We propose a feature learning algorithm to learn which system makes how much better predictions for what instances compared to the other. As the base of our learning framework, we have a randomized experiment, where there is an incumbent system and interventions are made randomly to overwrite the predictions of the incumbent with the predictions of a new system. We note the ``bandit feedback'' of the customers for all predictions in the experiment, some with the intervention and some without. For example, the incumbent system could be our current method of personalization, where we choose appropriate content based on user features using the current method, and the intervention is a newly developed method of personalization, which chooses the appropriate contents from the content pool based on user features in a different way. 

From the randomized experiment, we learn the feature representations which divide the population of instances into subpopulations where the difference in average customer feedback between those who were subjected to the intervention and those who were not, is statistically significantly different from $0$, with a level of significance $l$ and $l$ is a configurable parameter in our model. We use this information to derive the value of the intervention in consideration for each instance in the population based on its context, which we call derived personal valuation, depending on the membership of that particular instance in some subpopulation which exhibited statistically significant valuation, exclusivity of the subpopulation, the estimated average valuation for the subpopulation and its volatility. Note that even though in \cite{johansson:shalit:sontag:2016}, the authors used feature representations in the problem of counterfactual inference, our motivation for feature learning is completely different from them. In \cite{johansson:shalit:sontag:2016}, the authors used feature learning to reduce the selection bias in observational studies, whereas in this paper, we start with a randomized experiment which already removes the selection bias and we use feature learning to deduce conclusive results (signal strong compared to noise) for relatively smaller groups, which might be very different from the result at the global level (which could potentially still be inconclusive).

In our above example with two different personalization methods, we infer that the users with positive derived personal valuations (note that the derived personal valuation depends only on the context of the user in the form of user features like past browse and purchase activity) prefer the new method of personalization more than the current one. If the derived personal valuation is negative for everyone, we can safely discard the new method, as the current method universally dominates the new methods. Similarly, if all derived personal valuations are positive, the new method universally dominates the current method and we can safely replace the old method with the new one. However, a more realistic scenario is where the derived personal valuations range from negative to positive values, suggesting that for some users, the current method is better than the new one, whereas for some other users, it is the other way round. In the last and more realistic scenario, we might benefit in keeping the both the methods and build another layer of personalization in the system where based on user features, the system first decides on a method and then using the chosen method and the user features, picks content from the content pool for the user. 

Since our method of deriving personal valuation depends only on subpopulations where we have conclusive results (difference in average customer feedback is statistically significantly different from $0$, with a level of significance $l$), our derivation of personal valuation at a specified context  is more robust. In other words, our derived personal valuation is more free from the inherent noise in customer feedback.  The literature on contextual bandit problems  is dedicated to building robust estimators for each context (for example, see \cite{ li:chu:langford:schapire:2010, li:chu:langford:wang:2011, dudik:erhan:langford:li:2012}), but most of them are dedicated to the issue of imbalances in the observed data and the proposed solutions cleverly manage this imbalance. Since we start with a controlled experiment, such imbalances are not primary concern for us. However, our approach to derive robust personal valuations is focussed on the inherent noise in customer feedback and we attempt to make the estimator robust against this inherent noise. Thus our research is fundamentally different from the techniques used in the literature on the contextual bandit problem. To the best of our knowledge, no research has been devoted to the problem we addressed here.

With experiments, we show that the derived personal valuation for each instance could be leveraged in future to decide whether to intervene or not based on the features of the instance.

\begin{figure}
  \centering
  \includegraphics[width=\linewidth]{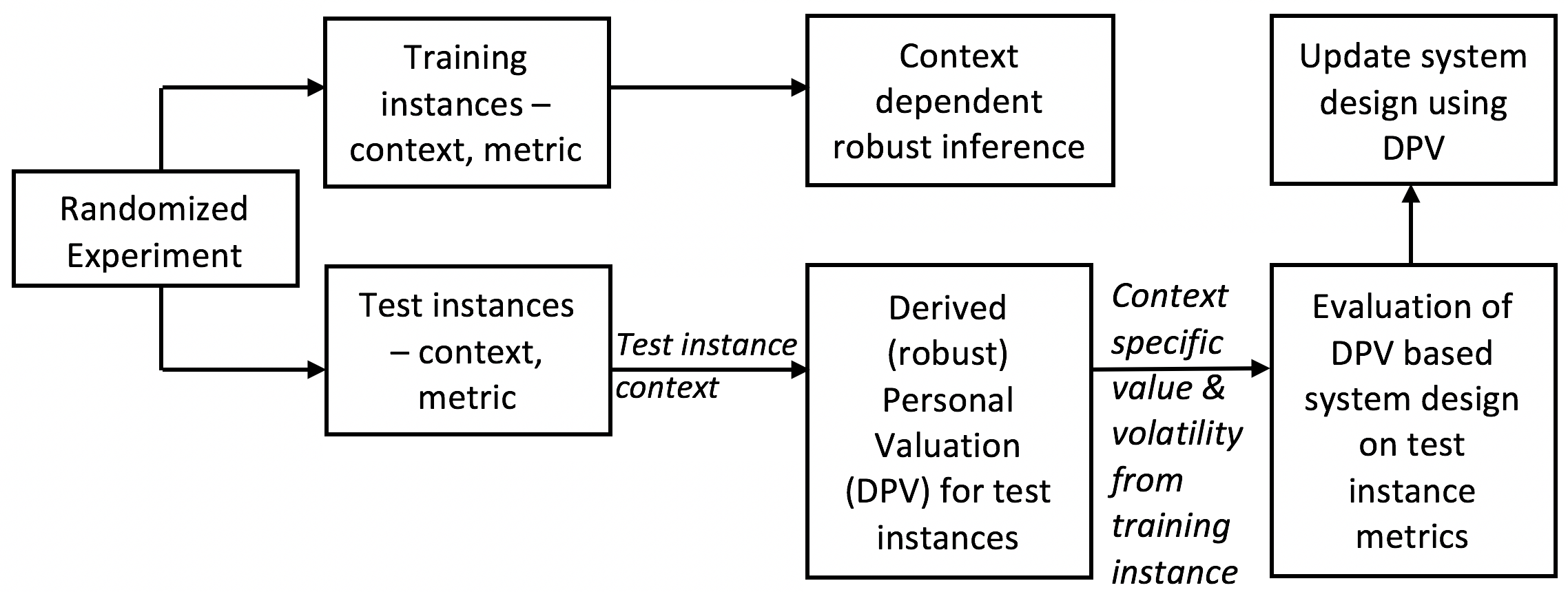}
  \caption{Application architecture design.}
  \label{fig:diagram2}
\end{figure}

Figure \ref{fig:diagram2} illustrates an example how the entire process would work. Suppose we have conducted a randomized experiment for 30 days with two versions of a webpage and collected customer clicks on the webpages. We want to understand which version of the webpage generates higher engagement or CTR (click-through-rate). We use, for example, the first 20 days for making context-based robust inferences. This is our set of training instances. We keep the last 10 days to evaluate how a derived personal valuation (DPV) based system design would perform. Note that this is how we will do a system design update based on DPV, where we learn from previous experiments (for example, the training instances) and use the learning to update system design that would impact future predictions (for example, the test instances). Suppose, no conclusion could be inferred at the global level using the training instances, in other words, CTR for version 1 is not statistically significantly different from CTR for version 2 with a level of significance, say 5\%. However, using gender of the user as a context gives us more information. Suppose, we find that women, in general, usually like version 2 better more than version 1 (based on CTR from training instances) and men prefer version 1 over version 2 and both these conclusions could be made with a level of significance, say 5\%. If this is our only conclusion, a derived personal valuation (DPV) based system design would suggest showing version 2 to women and version 1 to men. If women in the test instances indeed like version 2 more than version 1, and men in the test instances indeed like version 1 more than version 2, that would validate the fact that a DPV based system would work better than choosing either version1 or version 2 globally. In the second case where we choose one version globally, one of the group (either men or women) would be less engaged. Note that in making the inference that women like version 2 more, we have not used the test instances, but only the training instances. If we indeed used a DPV based system for the test instances, women would only see version 2 and we would not know how engaged they would be with version 1. Thus, for the validation of a DPV based system, it is necessary that the test instances are also part of the randomized experiment, so that we would have women seeing both version 1 and version 2 and would be able to compare the CTR difference without any selection bias and therefore, be able to understand how a DPV based system would perform. After this evaluation of the DPV-based system, if the DPV-based indeed works better, we would update our system using DPV, which in this example is showing version 2 to women and version 1 to men.


\section{Mathematical framework}
\label{sec:math_framework}

In this section, we provide the criterion for deciding whether a subpopulation exhibits a statistically significant valuation for the intervention in consideration, that is, the difference in average customer feedback is statistically significantly different from $0$, for predictions made with the intervention when compared to the ones made without the intervention, with a specified level of significance $l$ and we configure the parameter $l$ in our model. Then we construct an optimization problem we need to solve in order to identify subpopulations more likely to pass the criterion. Without loss of generality, we assume each feature could only take a finite number of real values. For features not directly satisfying this assumption, we appropriately merge values or discretize to satisfy this assumption.

To provide a mathematical framework for the problem, let us first introduce some notation. Let ${\bf{x}}_i = (x_{i,1},  x_{i, 2},  \cdots, x_{i, F})$ be the $i$-th instance (member of population) and $x_{i,j}$ be the value of the $j$-th feature of the $i$-th instance. In other words, ${\bf{x}}_i$ provides the context for instance $i$. Assume the total number of features used to represent context of an instance is $F$ and each instance assumes a value for each of the features, that is, a real value of $x_{i,j}$ is available for each instance $i$ and each feature $j$. Let $y_i$ be the metric or customer feedback for the $i$-th instance using which we are measuring valuation of the intervention in consideration, that is, if the intervention  impacts positively, the metric is expected to increase and if the intervention generates negative impact, the metric is expected to decrease. For example, a metric could be the number of clicks per page view. Usually the metric is driven by business goals. We could potentially consider several metrics together making $y_i$ a vector, but in this paper, we will restrict $y_i$ to be only a scalar. We assume the metric $y_i$ for each instance $i$ are mutually independent (in the probabilistic sense). In future references, we will call the customer feedback we chose to compare the system performances as the metric.

We assume a standard randomized experiment is set-up for the entire population. Thus the population is randomly divided into two groups, a test group $T$ and a control group $C$ and while the test group is subjected to predictions with the intervention, the control group is subjected to predictions without the intervention. Also, let us denote the whole population of instances as $\mathcal{P}$. Let ${\bf{x}}^T_i = (x^T_{i,1},  x^T_{i, 2},  \cdots, x^T_{i, F})$ be the feature vector for the $i$-th instance of the test group and ${\bf{x}}^C_j = (x^C_{j,1},  x^C_{j, 2},  \cdots, x^C_{j, F})$ be the feature vector of the $j$-th instance of the control group. Similarly, let $y^T_i$ be the metric or customer feedback (clicks, revenue etc.) for the $i$-th instance in the test group and $y^C_j$ be the metric or the customer feedback for the $j$-th instance in the control group.

\subsection{ Subpopulation eligibility}

First we establish a criterion for deciding whether a subpopulation shows a statistically significant impact of the intervention in consideration.  Since in the randomized experiment set-up, the membership in the test or control group is decided randomly independent of the context ${\bf{x}}$ of the instance, comparison of metrics for test and control groups restricted to any subpopulation based on ${\bf{x}}$, say the subpopulation $\{ \underline h({\bf{x}}) = \underline{v} \}$ where $\underline h(\cdot) : \mathbb{R}^F \to \mathbb{R}^K$ is a measurable function and $\underline v \in \mathbb{R}^K$, the measurement of the impact of the intervention in consideration for the subpopulation should also be free of any selection bias. In this paper, we restrict ourselves to linear feature representations and thus for us, $\underline h(\cdot) = {\bf{H}} \times \cdot$, where ${\bf{H}}$ is a $K \times F$ matrix.

We would like to identify subpopulations where there is a statistically significant impact of the intervention in consideration, or in other words, we want to find subpopulations where the difference in the average metrics for the test and control group restricted to the subpopulation is statistically significantly different from $0$, with a level of significance $l$. Using notation, the condition translates to finding ${\bf{H}}$ and $\underline v$ such that
\begin{align}\label{eq:feasibility}
& \frac{ \left|\left[ \bar y^T \big| {\bf{H}} \cdot {\bf{x}} = \underline v \right] - \left[ \bar y^C \big| {\bf{H}} \cdot {\bf{x}} = \underline v \right] \right|}{\sqrt{ \frac{Var\left[  y^T \big| {\bf{H}} \cdot {\bf{x}} = \underline v \right]}{\left| T \cap  \{{\bf{H}} \cdot {\bf{x}} = \underline v \}\right|} + \frac{Var\left[ y^C \big| {\bf{H}} \cdot {\bf{x}} = \underline v \right]}{\left| C \cap  \{{\bf{H}} \cdot {\bf{x}} = \underline v \}\right|} }} > q, \nonumber\\
 & \hskip 0.5 cm \Leftrightarrow \frac{\left| T \cap \{ {\bf{H}}\cdot {\bf{x}} = \underline v \} \right| \times \left| C \cap \{ {\bf{H}}\cdot {\bf{x}} = \underline v \} \right| }{\left| \{ {\bf{H}}\cdot {\bf{x}} = \underline v \} \right|} \nonumber \\
& \hskip 0.1 cm  \times  \frac{ {\left(\left[ \bar y^T \big| {\bf{H}}\cdot {\bf{x}} = \underline v \right] - \left[ \bar y^C \big| {\bf{H}}\cdot {\bf{x}} = \underline v \right] \right)}^2}{\frac{Var\left[  y^T \big| {\bf{H}}\cdot {\bf{x}} = \underline v \right] \cdot \left| C \cap  \{{\bf{H}}\cdot {\bf{x}} = \underline v \} \right| +  Var\left[ y^C \big| {\bf{H}}\cdot {\bf{x}} = \underline v \right] \cdot \left| T \cap  \{{\bf{H}}\cdot {\bf{x}} = \underline v \}\right|}{\left| \{ {\bf{H}}\cdot {\bf{x}} = \underline v \} \right|}} > q^2,
\end{align}
where $\left[\bar y^T \big| {\bf{H}} \cdot {\bf{x}} = \underline v \right]$ and $\left[ \bar y^C \big| {\bf{H}} \cdot {\bf{x}} = \underline v \right]$ are the average metrics from the test and control group respectively restricted to the subpopulation $\{ {\bf{H}} \cdot {\bf{x}} = \underline{v} \}$, $Var\left[y^T \big| {\bf{H}} \cdot {\bf{x}} = \underline v \right]$ and $Var\left[y^C\big| {\bf{H}} \cdot {\bf{x}} = \underline v \right]$ are the empirical variances of the metric in the test and control group respectively restricted to the subpopulation $\{ {\bf{H}} \cdot {\bf{x}} = \underline{v} \}$, $q^2$ is the $\left(1-l \right)$-th quantile of the distribution of the quantity on the LHS of \eqref{eq:feasibility} under the null hypothesis that the average metric difference between test and control group is $0$ (recall, $l$ is the level of significance of the test, so that under the null hypothesis, the probability of \eqref{eq:feasibility} being satisfied is exactly $l$) and the function $| \cdot |$ equals the size of the set in its argument. Under the null hypothesis that the intervention has no impact, the distribution of the quantity on the LHS of \eqref{eq:feasibility} could be approximated by the square of a variable following standard normal distribution.

\subsection{ Finding eligible subpopulations}

We call a subpopulation $\{ {\bf{H}}\cdot {\bf{x}} = \underline v \}$ {\it{eligible}} to be included in deriving personal valuation if it satisfies  \eqref{eq:feasibility}. Note that satisfying \eqref{eq:feasibility} is equivalent to rejecting a null hypothesis that the intervention has no impact, where the level of significance of the statistical test is $l$. We begin by noting that a subpopulation $\{ {\bf{H}}\cdot {\bf{x}} = \underline v \}$ may not be {\it{eligible}} according to \eqref{eq:feasibility} for one of the two reasons: either there is little or no impact of the intervention in consideration, or there is insufficient data to conclude anything, or both. These two reasons, in a way, complement each other. Note that if we have a lot of data, we could statistically significantly measure even tiny impacts and if we have a small amount of data, the impact needs to be huge in order for us to be able to measure that in a statistically significant way. On the other hand, if we have a huge impact of the intervention in consideration for the subpopulation, all we need is a very small amount of data to measure it in a statistically significant way, and if we have a tiny impact, we need huge amounts of data to do the same. Therefore, the reason for which a subpopulation is not {\it{eligible}}, whether it is the insufficiency of data or the relatively little impact of the intervention in consideration, is more of a subjective decision. 

If we could find out a way to identify the subpopulations which has the highest impact without actually checking \eqref{eq:feasibility}, we have satisfied our objective and we need not do anything more. However, no such method is known in its full generality and finding such a method seems to be a harder problem. Instead, we aim to identify subpopulations which have a lot of data, so that when we use \eqref{eq:feasibility} to measure the impact of intervention in consideration, even relatively smaller impacts could be measured in a  statistically significant way. 

We would focus on the first term on the LHS of \eqref{eq:feasibility}, which is  $ \frac{\left| T \cap \{ {\bf{H}}\cdot {\bf{x}} = \underline v \} \right| \times \left| C \cap \{ {\bf{H}}\cdot {\bf{x}} = \underline v \} \right| }{\left| \{ {\bf{H}}\cdot {\bf{x}} = \underline v \} \right|}$, and is a quantification of the amount of data for the subpopulation $\{ {\bf{H}}\cdot {\bf{x}} = \underline v\}$. This term could be re-written as $| \{ {\bf{H}}\cdot {\bf{x}} = \underline v\} |\cdot w^T\cdot \left(1 - w^T \right)$, where $w^T = \frac{\left| T \cap \{ {\bf{H}}\cdot {\bf{x}} = \underline v \} \right|}{| \{ {\bf{H}}\cdot {\bf{x}} = \underline v\} |}$ is the fraction of the subpopulation in the test group. Thus this term is dependent on the subpopulation size $| \{ {\bf{H}}\cdot {\bf{x}} = \underline v \} |$, as well as the fractions of the subpopulation in test and control group, given by $w^T$ and $\left(1 - w^T \right)$ respectively. We want find ${\bf{H}}$ so as to maximize the first quantity on the LHS of \eqref{eq:feasibility} for all subpopulations created by ${\bf{H}}$, viz. $\{ {\bf{H}}\cdot {\bf{x}} = \underline v_h \}$, $h = 1, 2, \cdots, H$. Since achieving that for all subpopulations created by ${\bf{H}}$ together might not be possible, we want to maximize the expected value of the quantity over all subpopulations created by ${\bf{H}}$. The expected value of the quantity over all subpopulations created by ${\bf{H}}$, where each subpopulation is weighted by its relative size simplifies to

\begin{align}\label{eq:objective_fn}
E_{\underline v_h} & \left[\frac{\left| T \cap \{ {\bf{H}}\cdot {\bf{x}} = \underline v_h \} \right| \times \left| C \cap \{ {\bf{H}}\cdot {\bf{x}} = \underline v_h \} \right| }{\left| \{ {\bf{H}}\cdot {\bf{x}} = \underline v_h \} \right|} \right] \nonumber \\
&=  \frac{1}{\left| \mathcal{P} \right|}  \sum_{i \in T} \sum_{j \in C} 1_{\{{\bf{H}}\cdot \left( {\bf{x}}^T_i - {\bf{x}}^C_j \right) = \underline 0 \}},
\end{align}
 where $1_C = 1$ if C is true, and $1_C = 0$ otherwise, are indicator functions indicating whether condition $C$ is true or not. Recall, the set $\mathcal{P}$ denotes the population of instances. The proof of the equivalence in \eqref{eq:objective_fn} is shown in Appendix \ref{appendix:objective_function_derivation}.
 
So, motivated by \eqref{eq:objective_fn}, we search for ${\bf{H}}$ which maximizes the RHS of \eqref{eq:objective_fn}. To formulate this as an optimization problem, we define a matrix ${\bf{Z}}$, whose columns are of the form $\left( {\bf{x}}^T_i - {\bf{x}}^C_j \right)$, where $i \in T$, and $j \in C$. Thus ${\bf{Z}}$ is a huge matrix with dimensions equal to $F \times \left|T \right|\cdot\left| C\right|$. Recall, $F$ is the total number of features describing instances in the population and $ \left|T \right|$ and $\left| C\right|$ are the sizes of the test and control group respectively. Let ${\bf{Z}}_{\cdot k}$ be the $k$-th column of ${\bf{Z}}$. Our optimization problem to search for ${\bf{H}}$ is formulated as 
\begin{align}\label{eq:opt_formulation}
\begin{array}{ccc}
 \displaystyle & \max_{{\bf{H}}, \left\{ a_k, k = 1, 2, \cdots, \left|T \right| \cdot \left| C \right| \right\}}   &{\sum_{k = 1}^{\left|T \right| \cdot \left| C \right|} a_k}  \\
\textrm{s.t.}  & a_k {\bf{H}} \cdot {\bf{Z}}_{\cdot k}    =  \underline 0,  & \forall k = 1, 2, \cdots, \left|T \right| \cdot \left| C \right|, \\
 & a_k  \in  \left\{ 0, 1\right\},  & \forall k = 1, 2, \cdots, \left|T \right| \cdot \left| C \right|. \\
\end{array}
\end{align}
Here the variables $a_k$ act like slack variables, in the sense that if ${\bf{H}} \cdot {\bf{Z}}_{\cdot k} \ne \underline 0$, then $a_k$ must be $0$ in order to satisfy the linear constraint in the optimization problem \eqref{eq:opt_formulation}. If ${\bf{H}} \cdot {\bf{Z}}_{\cdot k} = \underline 0$, the corresponding slack variable $a_k$ must assume the value $1$ in order to maximize the objective function $\sum_{k = 1}^{\left|T \right| \cdot \left| C \right|} a_k$ of the optimization problem \eqref{eq:opt_formulation}. Therefore, it is easy to see that if ${\bf{H}}, \left\{ a_k, k = 1, 2, \cdots, \left|T \right| \cdot \left| C \right| \right\}$ is a solution of the optimization problem \eqref{eq:opt_formulation}, then ${\bf{H}}$ and $\left\{ a_k, k = 1, 2, \cdots, \left|T \right| \cdot \left| C \right| \right\}$ will satisfy the condition $\sum_{k = 1}^{\left|T \right| \cdot \left| C \right|} a_k = \sum_{i \in T} \sum_{j \in C} 1_{\{{\bf{H}}\cdot \left( {\bf{x}}^T_i - {\bf{x}}^C_j \right) = \underline 0 \}}$, which is equal to $\left| \mathcal{P} \right|$ times the RHS of \eqref{eq:objective_fn}. So, it follows that ${\bf{H}}$ obtained as a solution from the optimization problem \eqref{eq:opt_formulation} would also maximize the LHS of \eqref{eq:objective_fn}, which is exactly what we wanted. 

Note that here, even if we restrict ourselves to linear feature representations to define the subpopulations, we do not aim to estimate a prediction function for $y_i$ based on ${\bf{x}}_i$, which is a major focus of methods trying to eliminate selection bias; see \cite{rosenbaum:rubin:1983, johansson:shalit:sontag:2016}.

\subsection{Reducing the search space of ${\bf{H}}$ matrices}

Our next goal is to reduce the search space of ${\bf{H}}$ by eliminating some redundancies and imposing some structure on ${\bf{H}}$ in the optimization problem \eqref{eq:opt_formulation}. We note that we could demand the rows of ${\bf{H}}$ to be orthonormal without changing the set of subpopulations we consider with the help of the following two propositions. The proofs follow from set equalities and are omitted here for space constraints.

\begin{prop}\label{prop:orthonormality}
The following two statements are true about the set of subpopulations generated by a matrix ${\bf{H}}$: 
\begin{enumerate}
\item If ${\bf{H}}$ is not full row rank, the set of subpopulations generated by ${\bf{H}}$, viz. $ \{ \{ {\bf{H}}\cdot  {\bf{x}} = \underline v_h \}, h = 1, 2, \cdots, H \}$, could also be generated by a lower dimensional matrix with lesser number of rows. 

\item If ${\bf{H}}$ is full row rank, the set of subpopulations generated by ${\bf{H}}$, viz. $ \{ \{ {\bf{H}}\cdot  {\bf{x}} = \underline v_h \}, h = 1, 2, \cdots, H \}$, could also be generated by a matrix with dimensions same as ${\bf{H}}$ and whose rows are orthonormal.
\end{enumerate}
\end{prop}

\subsection{Searching for multiple ${\bf{H}}$ matrices}

Our goal is to find as many subpopulations as possible, which satisfy \eqref{eq:feasibility}, or in other words, find as many subpopulations as possible, where the null hypothesis of no effect of intervention is rejected in the statistical test with level of significance $l$. The more such subpopulations we find, the more information we extract from the randomized experiment conducted. As discussed previously, the reason for satisfying \eqref{eq:feasibility} could be attributed to either the amount of data or the magnitude of the impact of the intervention inconsideration. In our optimization problem, we focussed on finding subpopulations with the most amount of data. Note that, while it is true that if the amount of data is very little, there is little chance for a subpopulation to satisfy \eqref{eq:feasibility}, but with reasonable amount of data, some subpopulations could still satisfy \eqref{eq:feasibility} if the magnitude of the impact from the intervention is high enough. It is highly likely that the entire population has the most data, but the impact of the intervention in consideration has no statistically significant impact for the entire population does not preclude the possibility of the impact of the intervention in consideration being statistically significant for a subpopulation with a lot less data.  

So, our search for feature representations does not end when we have a solution of optimization problem \eqref{eq:opt_formulation} and we keep looking for the next best one, which has less data than the previous one, but still could satisfy \eqref{eq:feasibility}. Here we discuss when we have found a set of matrices $\{ {\bf{H}}_1, {\bf{H}}_2, \cdots, {\bf{H}}_n \}$ and start searching for ${\bf{H}}_{\left(n + 1 \right)}$, what additional restrictions we can impose on the optimization problem \eqref{eq:opt_formulation} to search for the next best one. The following proposition suggests that the row space of ${\bf{H}}_{\left(n + 1 \right)}$ must not be a subset of the row space of ${\bf{H}}_i$ for $i = 1, 2, \cdots, n$. Let us denote the row space of ${\bf{H}}_i$ by $\mathcal{R} \left( {\bf{H}}_i \right)$. Once again, the proof follows from set equalities and is omitted here for space constraints.

\begin{prop}\label{prop: restriction_subsequent_matrices}
If $\mathcal{R} \left( {\bf{H}}_{\left(n + 1 \right)} \right) \subseteq \mathcal{R} \left( {\bf{H}}_i \right)$ for some $i = 1, 2, \cdots, n$, then the set of subpopulations generated by ${\bf{H}}_{\left(n + 1 \right)}$ is the same as the set of subpopulations generated by ${\bf{H}}_i$.
\end{prop}

In light of Proposition \ref{prop: restriction_subsequent_matrices}, we want to add the restriction that $\mathcal{R} \left( {\bf{H}}_{\left(n + 1 \right)} \right) \not \subseteq \mathcal{R} \left( {\bf{H}}_i \right)$ for ${\bf{H}}_i$, $i = 1, 2, \cdots, n$. The condition that $\mathcal{R} \left( {\bf{H}}_{\left(n + 1 \right)} \right) \not \subseteq \mathcal{R} \left( {\bf{H}}_i \right)$ could be re-written as the following condition: $\sum_{j = 1}^K {\bf{H}}_{\left(n + 1 \right), j \cdot} \left(\mathbb{I} - {\bf{H}}_i^T{\bf{H}}_i \right){\bf{H}}^T_{\left(n + 1 \right), j \cdot}> 0$, where ${\bf{H}}_{\left(n + 1 \right), j \cdot}$ is the $j$-th row of ${\bf{H}}_{\left( n + 1 \right)}$. The equivalence holds since ${\bf{H}}_i^T{\bf{H}}_i $ is the projection matrix for $\mathcal{R} \left( {\bf{H}}_i \right)$ and hence $\left(\mathbb{I} - {\bf{H}}_i^T{\bf{H}}_i \right)$ is an idempotent matrix. So, putting everything together, having found $ \{ {\bf{H}}_i, i = 1, 2, \cdots, n \}$, to find the $\left(n + 1 \right)$-th ${\bf{H}}$ matrix ${\bf{H}}_{\left(n + 1 \right)}$, we solve the following optimization problem:

\begin{align}\label{eq:opt_restricted_formulation_subsequent_steps}
\begin{array}{cl}
\displaystyle  \max_{{\bf{H}}_{\left(n + 1 \right)}, \left\{ a_k, k = 1, 2, \cdots, \left|T \right| \cdot \left| C \right| \right\}}   \sum_{k = 1}^{\left|T \right| \cdot \left| C \right|} a_k  & \\
 \textrm{s.t.}  \hskip 0.2 cm a_k {\bf{H}}_{\left(n + 1 \right)} \cdot {\bf{Z}}_{\cdot k}     =  \underline 0, & \forall k = 1, 2, \cdots, \left|T \right| \cdot \left| C \right|, \\
{\bf{H}}_{\left(n + 1 \right)} \cdot {\bf{H}}_{\left(n + 1 \right)}^{T}  =  \mathbf{I},   &\\
 \sum_{j = 1}^K {\bf{H}}_{\left(n + 1 \right), j \cdot} \left(\mathbb{I} - {\bf{H}}_i^T{\bf{H}}_i \right){\bf{H}}^T_{\left(n + 1 \right), j \cdot}  >  D, &  \forall i = 1, 2, \cdots, n,\\
  a_k   \in   \left\{ 0, 1\right\},   &  \forall k = 1, 2, \cdots, \left|T \right| \cdot \left| C \right|. \\
\end{array}
\end{align}
The parameter $D > 0$ could be chosen as an appropriate tuning parameter in the algorithm, which solves the optimization problem \eqref{eq:opt_restricted_formulation_subsequent_steps}. It is understood that when searching for the first matrix ${\bf{H}}_1$, that is $n = 0$, the fourth set of constraints in \eqref{eq:opt_restricted_formulation_subsequent_steps} will disappear.

Note that each run of the optimization problem gives us a feature representation ${\bf{H}}_{\left(n + 1 \right)}$ and with increasing $n$, the optimal value of the optimization problem \eqref{eq:opt_restricted_formulation_subsequent_steps} drops, indicating that expected quantity of data associated with  ${\bf{H}}_{\left(n + 1 \right)}$ (in the sense of \eqref{eq:objective_fn}) is reducing as $n$ increases. We stop when $n$ exceeds a preset threshold or the optimal value drops below a preset threshold. Note that the drop in optimal value of  ${\bf{H}}_{\left(n + 1 \right)}$ with increasing $n$ is not of concern, because while the drop means there is less expected quantity of data from the subpopulations generated by ${\bf{H}}_{\left(n + 1 \right)}$ (see \eqref{eq:objective_fn}), ultimately we want to identify all subpopulations which satisfy \eqref{eq:feasibility} and not only be restricted to subpopulations generated by ${\bf{H}}_{1}$. The more subpopulations we find that satisfy  \eqref{eq:feasibility}, the more information we extract from our randomized experiment.

Note that in earlier discussion, we fixed the dimension of ${\bf{H}}_{\left(n + 1 \right)}$ as $K \times F$, where $F$ is the total number of features defining a context ${\bf{x}}$. While we cannot change $F$ as it is given to us, we do have some flexibility in the choice of $K$. Instead of fixing a particular $K$, we could start from $K = 1$ (dimension of ${\bf{H}}_{\left(n + 1 \right)}$ is $K \times F$ ) and then continue increasing $K$, thus increasing the granularity of the subpopulations. Note that the required magnitude of impact in order to satisfy \eqref{eq:feasibility} goes up as a consequence, which in turn reduces the likelihood of condition \eqref{eq:feasibility} being satisfied for those subpopulations. Thus, it is advisable to keep the $K$ much lower compared to $F$. By following this, in our final set of identified subpopulations that satisfy \eqref{eq:feasibility}, some could be generated by ${\bf{H}}_{\left(n + 1 \right)}$-s of dimensions $K_1 \times F$ and some could be generated by ${\bf{H}}_{\left(n + 1 \right)}$-s of dimensions $K_2 \times F$, where $K_1 \ne K_2$. Note that our only aim is to identify as many subpopulations as possible that satisfy \eqref{eq:feasibility} and we do not care whether they are characterized by ${\bf{H}}$ matrices of the same dimensions or not.

However, there might be computational limitations as to how long we can prolong our search of ${\bf{H}}$ matrices and after a while, the expected quantity of data (in the sense of  \eqref{eq:objective_fn} ) associated with an  ${\bf{H}}$ matrix will become very low. This, in turn, would result in the subpopulations generated by those  ${\bf{H}}$ matrices having lesser and lesser data, which means those subpopulations are more and more unlikely to satisfy \eqref{eq:feasibility}. Thus, we stop when we reach our computational limit or when the expected quantity of data (in the sense of  \eqref{eq:objective_fn} ) associated with an  ${\bf{H}}$ is low. Even though with this stopping condition, we may have missed some subpopulation which could have satisfied  \eqref{eq:feasibility}, in the process, we have extracted a lot more information from the randomized experiment than just the comparison of the average metrics of the test and control group at the global level.

\section{Algorithm to find subpopulations}
\label{sec:algorithm}

To solve the optimization problem \eqref{eq:opt_restricted_formulation_subsequent_steps}, we consider a Lagrangian relaxation of the problem given by 

\begin{align}\label{eq:opt_restricted_formulation_subsequent_steps_lagrangian}
\displaystyle \max_{{\bf{H}}_{\left( n + 1\right)}, \left\{ a_k \in \{0, 1\}, k = 1, 2, \cdots, \left|T \right| \cdot \left| C \right| \right\}}  L = \sum_{k = 1}^{\left|T \right| \cdot \left| C \right|} a_k + & \nonumber \\
 \sum_{j = 1}^ K \sum_{k = 1}^{\left|T \right| \cdot \left| C \right|} \lambda_{j, k} a_k {\bf{H}}_{\left( n + 1\right), j \cdot}  \cdot {\bf{Z}}_{\cdot k} + &\nonumber\\
 \hskip 1 cm  \sum_{i = 1}^n \mu_i \left( \sum_{j = 1}^K {\bf{H}}_{\left(n + 1 \right), j \cdot} \left(\mathbb{I} - {\bf{H}}_i^T{\bf{H}}_i \right){\bf{H}}^T_{\left(n + 1 \right), j \cdot} - D \right), &\nonumber\\
 s. t. \hskip 0.2 cm  {\bf{H}}_{\left(n + 1 \right)} \cdot {\bf{H}}_{\left(n + 1 \right)}^{T} =  \mathbf{I}, &
\end{align} 
where ${\bf{H}}_{\left( n + 1\right), j \cdot}$ is the $j$-th row of ${\bf{H}}_{\left( n + 1\right)}$ and  $\mu_i$ and $\lambda_{j, k}$ are penalty constants; see \cite{nocedal:wright:2006}. We take a greedy approach and solve \eqref{eq:opt_restricted_formulation_subsequent_steps_lagrangian} by updating ${\bf{H}}_{\left( n + 1\right)}$ and $\left\{ a_k \in \{0, 1\}, k = 1, 2, \cdots, \left|T \right| \cdot \left| C \right| \right\}$ in sequence. We choose $\mu_i = \frac{1}{2}$ for all $i = 1, 2, \cdots, n$, and at each update, we change the constant $\lambda_{j, k}$ as $\lambda_{j, k} = -\text{sign} \left( {\bf{H}}_{\left( n + 1\right), j \cdot} \cdot {\bf{Z}}_{\cdot k} \right)$, where ${\bf{H}}_{\left( n + 1\right)}$ in the last step is used for computation of the constant. 

We update ${\bf{H}}_{(n+1)}$ by gradient descent, where we move the ${\bf{H}}_{(n+1)}$ slightly in the direction of the derivative of $L$ given in \eqref{eq:opt_restricted_formulation_subsequent_steps_lagrangian} w.r.t. ${\bf{H}}_{(n+1)}$. Also, leveraging Proposition \ref{prop: restriction_subsequent_matrices}, we could claim that it is good enough to only consider the update vector projected in the orthogonal space of the row space of current ${\bf{H}}_{(n+1)}$. So, finally, the update to the matrix ${\bf{H}}_{\left(n + 1 \right)}$ would be: for a small $\epsilon > 0$, 
\begin{align}\label{eq:Hn_update}
&{\bf{H}}_{\left( n + 1\right)}^{updated} =  {\bf{H}}_{\left( n + 1\right)} + \epsilon \frac{\frac{\partial L}{\partial {\bf{H}}_{\left( n + 1\right)}}\cdot \left( \mathbb{I} - {\bf{H}}_{\left(n + 1\right)}^T{\bf{H}}_{\left(n + 1\right)} \right)}{\left|\left|\frac{\partial L}{\partial {\bf{H}}_{\left( n + 1\right)}} \cdot \left( \mathbb{I} - {\bf{H}}_{\left(n + 1\right)}^T{\bf{H}}_{\left(n + 1\right)} \right) \right| \right|}.
\end{align}

The next step is updating $\left\{ a_k \in \{0, 1\}, k = 1, 2, \cdots, \left|T \right| \cdot \left| C \right| \right\}$. To do that, first we compute $MAX = \max_{k = 1}^{\left|T \right| \cdot \left| C \right|} \max_{i = 1}^K  \left| {\bf{H}}_{\left(n + 1 \right), i \cdot}  \cdot {\bf{Z}}_{\cdot k} \right|$. Then we update each $a_k$ in the following way: if the condition $\max_{i = 1}^K  \left| {\bf{H}}^{updated}_{\left(n + 1 \right),i \cdot}  \cdot {\bf{Z}}_{\cdot k} \right| > \theta \cdot MAX$ is satisfied for some value of $0 < \theta < 1$, we set $a_k = 0$ and otherwise, we set $a_k = 1$. The parameter $\theta$ could be tuned for the speed of convergence of the algorithm. See Algorithm \ref{alg:nth_matrix_search} below for more execution details. 

Now we select initializations of the variables. Note that the optimal value of the slack variable $a_k$ takes the value $1$ if and only if the features of the corresponding pair are equal in value once premultiplied by ${\bf{H}}_{\left(n + 1 \right)}$. We hope that for any initial choice of $\left\{ a_k \in \{0, 1\}, k = 1, 2, \cdots, \left|T \right| \cdot \left| C \right| \right\}$,  the appropriate ${\bf{H}}_{\left(n + 1 \right)}$ would be able to do that for every pair and choose $a_k = 1$ for $k = 1, 2, \cdots, \left|T \right| \cdot \left| C \right|$. For the initial choice of ${\bf{H}}_{\left(n + 1 \right)}$, we will perturb the last found solution ${\bf{H}}_n$ a little as shown below:
\begin{align}\label{eq:Hn_initial}
&{\bf{H}}_{\left( n + 1\right)}^{initial} = {\bf{H}}_n + \epsilon \frac{\frac{\partial L}{\partial {\bf{H}}_{n}}\cdot \left( \mathbb{I} - {\bf{H}}_{n}^T{\bf{H}}_{n} \right)}{\left|\left|\frac{\partial L}{\partial {\bf{H}}_{n}} \cdot \left( \mathbb{I} - {\bf{H}}_{n}^T{\bf{H}}_{n} \right) \right| \right|}.
\end{align}

Note that ${\bf{H}}_{n}$ is the optimal solution for the optimization problem \eqref{eq:opt_restricted_formulation_subsequent_steps} with $n$ replaced by $(n-1)$. So, ${\bf{H}}_{n}$ satisfies all the constraints on ${\bf{H}}_{\left(n + 1 \right)}$ except for the additional constraint imposed when $n$ is incremented by 1 in the optimization problem \eqref{eq:opt_restricted_formulation_subsequent_steps}, that is, ${\bf{H}}_{\left(n + 1 \right)}$ cannot belong to $\mathcal{R} \left( {\bf{H}}_n \right)$. So, we hope that the perturbation in \eqref{eq:Hn_initial} will satisfy all constraints on  ${\bf{H}}_{\left(n + 1 \right)}$. For initialization of ${\bf{H}}_1$, start with a $K \times K$ identity matrix appended by a zero matrix of dimensions $K \times (F - K)$. 

\begin{algorithm}
\caption{Algorithm to select ${\bf{H}}_{\left(n + 1 \right)}$}
\label{alg:nth_matrix_search}
\begin{algorithmic}[1]
\Procedure{Search for ${\bf{H}}_{\left(n + 1 \right)}$}{ Start with ${\bf{H}}_{\left( n + 1\right)} = {\bf{H}}_{\left( n + 1\right)}^{initial}$ as in \eqref{eq:Hn_initial}.} \\
Start with $a_k = 1$ $\forall k = 1, 2, \cdots, |T|\cdot|C|$.
\\Compute $ \Delta_{i \cdot} = \sum_{k = 1}^{\left|T \right| \cdot \left| C \right|} -\text{sign} \left( {\bf{H}}_{\left( n + 1\right), i \cdot}  \cdot {\bf{Z}}_{\cdot k} \right) a_k {\bf{Z}}_{\cdot k}^T$ $\forall i = 1, 2, \cdots, K$ and $MAX = \max_{k = 1}^{\left|T \right| \cdot \left| C \right|} \max_{i = 1}^K  \left| {\bf{H}}_{ \left( n + 1\right), i \cdot}  \cdot a_k{\bf{Z}}_{\cdot k} \right|$.
\\If $MAX < \gamma$, STOP.
\\If $\frac{\partial L}{\partial {\bf{H}}_{\left(n + 1 \right)}} \cdot \left(\mathbb{I} - {\bf{H}}^T_{\left(n + 1 \right)} {\bf{H}}_{\left(n + 1 \right)} \right) = \left[ \Delta + {\bf{H}}_{\left( n + 1\right)} \left( \sum_{i = 1}^n \left( \mathbb{I} - {\bf{H}}_i^T{\bf{H}}_i \right)\right)\right] \left( \mathbb{I} - {\bf{H}}_{\left(n + 1\right)}^T{\bf{H}}_{\left(n + 1\right)} \right) \ne \underline 0$, update ${\bf{H}}_{\left(n + 1 \right)}$ as in \eqref{eq:Hn_update}, else if $a_k$-s are updated at least once, STOP, else try a different initial ${\bf{H}}_{\left(n + 1 \right)}$, say, by changing $\epsilon$ in \eqref{eq:Hn_initial}.
\\Orthonormalize rows of ${\bf{H}}_{\left(n + 1 \right)}$ following Gram-Schmidt algorithm.
\\Set $a_k = 0$ if $\max_{i = 1}^K  \left| {\bf{H}}_{\left(n + 1 \right), i \cdot}  \cdot {\bf{Z}}_{\cdot k} \right| > \theta \cdot MAX$, otherwise, set $a_k = 1$.
\\Go back to step 3.
\EndProcedure
\end{algorithmic}
\end{algorithm}

\section{Measuring personal valuation for each instance}
\label{sec:personal_valuation}

The subpopulations we have identified in the previous section might overlap with each other and in this section, we focus on deriving the valuation of the impact for each instance. An instance might be part of several subpopulations which could potentially have different verdicts on the benefits of the intervention. Some subpopulation that the instance is part of (based on its context), may have a negative effect of the intervention, while some other subpopulation that it is part of, has a positive effect. Thus, given an instance with its context, we need to determine whether intervening with give us better feedback or not. This is what we focus on here.

We assume through the procedures described in the previous sections, we have found a set of subpopulations $\mathcal{S}$ of the form $\mathcal{S} = \{ {\bf{H}} \cdot {\bf{x}}  = \underline v\}$ which satisfy \eqref{eq:feasibility}. We derive the valuation of an instance with context ${\bf{x}}$ as follows:
\begin{align}\label{eq:personal_valuation}
v\left( {\bf{x}} \right) &= \sum_{ \{ \mathcal{S}:  {\bf{x}} \in \mathcal{S}\text{ and } \mathcal{S} \text{ satisfies \eqref{eq:feasibility}} \}}w_{\mathcal{S}, {\bf{x}}}v\left( \mathcal S \right),
\end{align}
where the weights $w_{\mathcal{S}, {\bf{x}}}$ are described  below and $v\left( \mathcal S \right)$ is the average valuation for the subpopulation $\mathcal{S}$ as found from the randomized experiment given by $v\left( \mathcal S \right) =  \left[ \bar y^T \big| \mathcal S \right] - \left[ \bar y^C \big| \mathcal S \right]$
where $\left[ \bar y^T \big| \mathcal S \right]$ and $\left[ \bar y^C \big| \mathcal S \right]$ are the average metrics from the test and control group respectively restricted to the subpopulation $\mathcal{S}$. Also, note that if no subpopulation satisfies the condition in the sum on the RHS of \eqref{eq:personal_valuation}, the value $v\left( {\bf{x}} \right)$ is the empty sum, which is $0$.

Intuitively, the weights  $w_{\mathcal{S}, {\bf{x}}}$ should have an inverse relationship with the volatility of the average metric $v\left( \mathcal S \right)$, as higher volatility means less confidence in our estimate of the average valuation $v\left( \mathcal S \right)$ for the subpopulation $\mathcal{S}$. Also, the weights  $w_{\mathcal{S}, {\bf{x}}}$ should penalize bigger subpopulations as they reduce the volatility of $v\left( \mathcal S \right)$ by adding more data and thus the individual valuations of its members (members of the subpopulation $\mathcal{S}$) are not necessarily close to the average valuation $v\left( \mathcal S \right)$ of the subpopulation $\mathcal S$. Note that now we are only interested in deriving the valuation of the instance with context ${\bf{x}}$ and not about the average valuation for a subpopulation that it belongs to. Including all these intuitions, we compute the weights by solving the following set of equations:
\begin{align}\label{eq:weight_proportionality}
\sum_{\{ \mathcal{S}:  {\bf{x}} \in \mathcal{S}\text{ and } \mathcal{S} \text{ satisfies \eqref{eq:feasibility}} \} } w_{\mathcal{S}, {\bf{x}}} = 1,  \hskip 1 cm w_{\mathcal{S}, {\bf{x}}} \propto  \frac{1}{ \sigma \left[ v\left( \mathcal{S} \right) \right] } \sqrt{\frac{1}{ \left| \mathcal{S} \right|}},
\end{align}
where $\sigma \left[ v\left( \mathcal{S} \right) \right] $ is the volatility of $v\left( \mathcal{S} \right) $. Note that the instance represented by its context vector ${\bf{x}}$ plays a role in defining the weights through \eqref{eq:personal_valuation}, where the summands are determined by ${\bf{x}}$.

Note that we could simplify the term that the weights $w_{\mathcal{S}, {\bf{x}}}$ given in \eqref{eq:weight_proportionality} are inversely proportional to, as 
\begin{align}\label{eq:weight_simplification}
\sigma \left[ v\left( \mathcal{S} \right) \right] & \cdot \sqrt{ \left| \mathcal{S} \right |} = \sqrt{\frac{Var\left[  y^T \big| \mathcal{S} \right]}{\left| T \cap  \mathcal{S} \right|} + \frac{Var\left[ y^C \big| \mathcal{S}  \right]}{\left| C \cap \mathcal{S} \right |} } \cdot \sqrt{ \left| \mathcal{S} \right |} \nonumber \\
&= \sqrt{\frac{\left| C \cap \mathcal{S} \right |}{ \left| \mathcal{S} \right| } \cdot Var\left[  y^T \big| \mathcal{S} \right] + \frac{\left| T \cap  \mathcal{S} \right|}{ \left| \mathcal{S} \right| } \cdot Var\left[ y^C \big| \mathcal{S}  \right]  } \nonumber\\
& \hskip 1 cm \times  {\left( \frac{\left| T \cap  \mathcal{S} \right|}{\left| \mathcal{S} \right |} \cdot  \frac{\left| C \cap  \mathcal{S} \right|}{\left| \mathcal{S} \right |} \right)}^{-\frac{1}{2}}.
\end{align}
In a controlled experiment as is our base set-up, we could assume the second term on the RHS of  \eqref{eq:weight_simplification} to be close to a constant for any reasonably large subpopulation and subpopulations need to be reasonably large to be {\textit{eligible}} according to \eqref{eq:feasibility}. Thus, the weights are dominated by the first term on the RHS of \eqref{eq:weight_simplification}, which means subpopulations where the metrics (customer feedbacks) are less volatile, will get higher weights than those with higher volatility, which conforms with the intuition that we trust those subpopulations more where the metrics are more consistent.

We potentially could derive personal valuation using methods in \cite{johansson:shalit:sontag:2016}, even though it was not the primary objective of the paper. However, note that the authors in \cite{johansson:shalit:sontag:2016} have used representation learning in removing selection bias, whereas our basic set-up is a randomized experiment and therefore, we do not have any selection bias to begin with. Blindly applying methods of \cite{johansson:shalit:sontag:2016} on the results of a randomized experiment would result in unnecessary overfitting. Moreover, the methods described in \cite{johansson:shalit:sontag:2016} would require a known form of prediction function (in  \cite{johansson:shalit:sontag:2016}, the authors optimize within a family of prediction functions), whereas we proceed without any assumption on the prediction function and do not even need one for our modeling.

Note that our derivation of personal valuation is based on {\textit{eligible}} subpopulations (see \eqref{eq:feasibility}) we found from the randomized experiment. Thus, in deriving the personal valuation at context ${\bf{x}}$, we automatically discard those subpopulations which contain ${\bf{x}}$, but where the first order difference (difference in mean) do not rise above the second-order noise (standard deviation), making the derived personal valuation (DPV) at context ${\bf{x}}$ more robust. In the literature on contextual bandit problems (for example, see \cite{ li:chu:langford:schapire:2010, li:chu:langford:wang:2011, dudik:erhan:langford:li:2012}), research has been carried out to reduce bias and variance of estimators for each context ${\bf{x}}$, but most of them are dedicated to the issue of imbalance in the training data and the proposed solutions cleverly manage this imbalance. Since our base set-up is a controlled experiment, such imbalances are not primary concern for us. On the other hand, even with the controlled experiment, our conclusions are crippled by the inherent noise in the customer feedbacks, which we attempt to resolve. Thus our approach to derive robust personal valuations is fundamentally different from the techniques used in the literature on contextual bandit problem.

\section{Applications}
\label{sec:applications}

One of the applications is finding relevant target population for a similar future intervention. Given the features for an instance, we can compute derived personal valuations (DPV) even if the instance was not part of the randomized experiment used to identify {\it{eligible}} subpopulations using \eqref{eq:feasibility}. We can assume that a subpopulation with relatively higher DPV will provide more incremental metrics in future interventions than a subpopulation with relatively lower DPV. For example, if the intervention is a new method of personalization as opposed to an existing one, we could show personalized content according to the new method for those who have the positive DPV for this intervention and keep running the current method for the rest. The scope of this  application is well beyond personalization, for example, one could use this to target audience for an online ad campaign, where the learning framework is used on a previous similar ad campaign and the intervention is an ad, as opposed to not running any ad campaign at all. In this application, we can target populations with highest DPV to best utilize campaign costs on a receptive audience.

The second application is identifying scope for improvement in the intervention in consideration. The groups with low or negative DPV for a given intervention such as a new method for personalization represent the population for which intervention did not perform well. Thus improvement of the new method of personalization can focus on such groups. Alternatively future interventions can choose to exclude such groups to optimize benefits as suggested before.

\subsection{Validation framework}

For empirical validation of how a DPV-based system design would work, we propose the following validation framework. We run a randomized experiment where predictions from a system, say, system A, is used as intervention and predictions from another system, say system B, is used as the default option. Customer feedback is collected on all predictions from the experiment. In the validation framework, we divide the randomized experiment data into training and test data instances. We select first 80\% of all instances in chronological order as training data and the remaining as test data. We use training data to identify {\it{eligible}} subpopulations which satisfy \eqref{eq:feasibility} and then use them to compute the DPV for instances in the test data. For each metric, we divide test instances in multiple groups in the order of their DPV by categorizing based on quartiles of DPV, everyone below quartile 1 is one group ($<$Q1), everyone below median but above quartile 1 is another group (Q1- Q2) and so on... Now for each DPV based group in test instances, we note the difference in average metric from those subjected to the intervention and those who were not, and call that our average incremental metric for the group. If DPV derived for the test instances are actually indicative of how each system will perform compared to the other, we expect to see increasing average incremental metric with increasing DPV. Thus the groups with higher DPV would have more incremental metric than groups with lower DPV. Since the training and test group are separated in a chronological order, this is exactly how we could use DPV in system design, where we learn from our past experiments which system works better for which instances, and use the DPV of future instances to decide the best system for the instance thus optimizing overall performance. Recall that in deriving the DPV for the test instances, we only used the metrics from training instances and the context for test instances, but never used the metrics for the test instances.  As shown in Table \ref{click-metric-table}, in our empirical experiment, DPV for the test instances were indeed indicative of system preference of the test instances.

\subsection{Results}

For our empirical experiment, the intervention was Whole Page Personalization (personalizing different modules of the page together) as opposed to separately personalizing different modules of the webpage. We ran a randomized experiment for 22 days, where a randomly selected fraction of online users were part of the experiment. The users in the randomized experiment were randomly divided into test and control group. The test group was exposed to Whole Page personalization and the control group saw independently personalized modules on the same webpage. We considered click through rate in a particular module category for metric within the web session. The metric assumes value $0$ if there was no click for given page view of a module category. We considered several other metrics, where each metric corresponds to clicks restricted to one category of content. We used past site activities of the user in different categories as context/features which are used to characterize an user at the time of the webpage visit (an instance is a user at the time of the webpage visit in this experiment).

We used first 16 days of the randomized experiment as training data and last 6 days as the test data, as suggested in our validation framework. We identified subpopulations satisfying \eqref{eq:feasibility} from training data and used them to derive personal valuations for users in the test data. Note that the features are determined by user and time-dependent user activity-based features. So, even if the same user comes twice or more during the randomized experiment, they will be treated as different instances. For the experiment below, we fix the level of significance at 30\% (recall that the level of significance $l$ is a configurable parameter in our model).

We present the results for two metrics on the test instances (Table \ref{click-metric-table}): CTR restricted to Category A and Category B modules. For both Category A and B we see that the average CTR difference between test and control groups increased significantly as the DPV increased for the groups. This means that Whole Page Personalization impact increases with increased DPV for the test instances. We only report the results for those DPV-based groups for whom the difference in CTR (restricted to the category) between those receiving Whole Page Personalization and those who did not, was statistically significantly different from $0$ with level of significance 30\%. Note that due to limitation of data in our test instances, all DPV-based groups in the test instances might not produce conclusive results about system preferences. However, wherever they do, they show DPV being indicative of system preferences.

From the training data (first 16 days of the experiment), the category A CTR (only clicks on content of category A is considered) difference between those who received Whole Page Personalization and those who did not, is -0.13\% with standard deviation of 0.23\%.  So, note that category A CTR is not statistically significantly different for the two groups at a global level. So, in a standard set-up, we would conclude there is no difference in these two experiences in terms of category A CTR metric. However, we found smaller subpopulations with conclusive preferences (restricted to the subpopulations, difference in category A CTR between those who received Whole Page Personalization and those who did not, were statistically significantly different from $0$) and using those, we derived the personal valuations. As results in Table \ref{click-metric-table} illustrate, with category A CTR as the chosen metric, a DPV-based system design would extract much more information from the experiment.

From the training data (first 16 days of the experiment), the category B CTR (only clicks on content of category B is considered) difference between those who received Whole Page Personalization and those who did not, is 0.34\% with standard deviation of 0.30\%.  So, note that category B CTR is statistically significantly different for the two groups at a global level. So, in a standard set-up, we would conclude Whole Page personalization is better in terms of category B CTR metric and impose that on everyone (assuming we are only interested in category B CTR). However, as results in Table \ref{click-metric-table} illustrate, we found a DPV-based group ($<$Q1) which actually does not prefer Whole Page Personalization (as measured by category B CTR). In this case as well, we would benefit from a DPV based system design.

\begin{table}[t]
  \caption{Table for metric - click through rate (CTR)}
  \label{click-metric-table}
  \centering
  \begin{tabular}{lllll}
    \toprule
     && Difference & Standard  &  DPV-\\ 
        &  DPV                 & in & deviation of &  based\\ 
           & based & average & difference in & group \\
    Metric & groups & CTR &  average CTR & size \\
    \midrule
    Category A CTR & $<$Q1 & 0.058 & 0.043 & 160  \\
    Category A CTR &Q2-Q3 & 0.135 & 0.108 & 87   \\
    Category B CTR & $<$Q1 & -0.032 & 0.030 & 156 \\
    Category B CTR & Q1-Q2 & 0.079 & 0.047 & 163  \\
    \bottomrule
  \end{tabular}
\end{table}

\section{Conclusion}
\label{sec:conclusion}

We have proposed a feature learning algorithm to identify optimal system for a given instance based on its context. We have shown that our learning could be leveraged to target populations for future interventions as well as personalize the choice of optimal systems. A framework that leverages such personal preference of optimal systems will generate prediction in a two-layered approach: first choose the preferred system and then make the appropriate prediction using the preferred system. Further research is needed to get clarity on how personalized the choice of systems can be, how many systems a framework can support etc. to build a framework that uses the personalized choice of systems at scale. While we proposed one greedy approach to solve the optimization problem \eqref{eq:opt_restricted_formulation_subsequent_steps}, further research needs to explore other possibly better ways of solving the optimization problem. Further research could also drive down the computational time. It would also be fruitful to invest in research to estimate noise for DPV and intelligent use of it in system design.

\section{Acknowledgement}

We sincerely thank Shyam Rapaka, who actively contributed in the material presented in this paper during his tenure at Walmart Labs.

\bibliographystyle{ACM-Reference-Format}
\bibliography{sample-bibliography.bib}

\appendix
\section{Derivation of objective function used in optimization problem formulation} \label{appendix:objective_function_derivation}
The expected value of the amount of data over all subpopulations created by ${\bf{H}}$, where each subpopulation is weighted by its relative size simplifies to
\begin{align}
E_{\underline v_h}& \left[\frac{\left| T \cap \{ {\bf{H}}\cdot {\bf{x}} = \underline v_h \} \right| \times \left| C \cap \{ {\bf{H}}\cdot {\bf{x}} = \underline v_h \} \right| }{\left| \{ {\bf{H}}\cdot {\bf{x}} = \underline v_h \} \right|} \right] \nonumber \\
&= \sum_{h = 1}^H \frac{\left| \{ {\bf{H}}\cdot {\bf{x}} = \underline v_h \} \right|}{\left| \mathcal{P} \right|} \cdot \frac{\left| T \cap \{ {\bf{H}}\cdot {\bf{x}} = \underline v_h \} \right| \times \left| C \cap \{ {\bf{H}}\cdot {\bf{x}} = \underline v_h \} \right| }{\left| \{ {\bf{H}}\cdot {\bf{x}} = \underline v_h \} \right|} \nonumber \\
&= \frac{1}{\left| \mathcal{P} \right|}  \sum_{h = 1}^H \left(\sum_{i \in T} 1_{\{{\bf{H}}\cdot {\bf{x}}^T_i = \underline v_h \}} \right) \times \left(\sum_{j \in C} 1_{\{{\bf{H}}\cdot {\bf{x}}^C_j = \underline v_h \}} \right), \nonumber \\
&=   \frac{1}{\left| \mathcal{P} \right|}  \sum_{h = 1}^H \sum_{i \in T} \sum_{j \in C} 1_{\{{\bf{H}}\cdot {\bf{x}}^T_i = \underline v_h \}} \times 1_{\{{\bf{H}}\cdot {\bf{x}}^C_j = \underline v_h \}}, \nonumber \\
&= \frac{1}{\left| \mathcal{P} \right|}  \sum_{i \in T} \sum_{j \in C}  \sum_{h = 1}^H  1_{\{{\bf{H}}\cdot {\bf{x}}^T_i = \underline v_h \}} \times 1_{\{{\bf{H}}\cdot {\bf{x}}^C_j = \underline v_h \}}, \nonumber \\
&= \frac{1}{\left| \mathcal{P} \right|}  \sum_{i \in T} \sum_{j \in C}  1_{\{{\bf{H}}\cdot {\bf{x}}^T_i = {\bf{H}}\cdot {\bf{x}}^C_j \}}, \nonumber \\
&=  \frac{1}{\left| \mathcal{P} \right|}  \sum_{i \in T} \sum_{j \in C} 1_{\{{\bf{H}}\cdot \left( {\bf{x}}^T_i - {\bf{x}}^C_j \right) = \underline 0 \}},
\end{align}
where $1_C = 1$ if C is true, and $1_C = 0$ otherwise, are indicator functions indicating whether condition $C$ is true or not. Recall, the set $\mathcal{P}$ denotes the population of instances.

\end{document}